# A Model-Based Approach to Predicting Predator-Prey & Friend-Foe Relationships in Ant Colonies

Karthik Narayanaswami
*College of Computing, Georgia Institute of Technology*
*Karthik.narayanaswami@cc.gatech.edu*

**Abstract**

Understanding predator-prey relationships among insects is a challenging task in the domain of insect-colony research. This is due to several factors involved, such as determining whether a particular behavior is the result of a predator-prey interaction, a friend-foe interaction or another kind of interaction. In this paper, we analyze a series of predator-prey and friend-foe interactions in two colonies of carpenter ants to better understand and predict such behavior. Using the data gathered, we have also come up with a preliminary model for predicting such behavior under the specific conditions the experiment was conducted in. In this paper, we present the results of our data analysis as well as an overview of the processes involved.

## 1. Introduction

The study of social insects for the purposes of computational research has always been a domain of interest. To this end, several tools and technologies have been developed with the explicit purpose of observing, analyzing and predicting the behavior of such insect colonies.

However, one of the challenges of this domain has been in effectively understanding the causality behind the various observed behaviors. There are several areas of interest within the domain of social insect behavior where working, robust models have been developed that accurately predict certain behavior, while there are several other domains where this has not quite been the case.

One such domain is the study and analysis of predator-prey and friend-foe behaviors in ants. While there has been a lot of biological work in this particular domain, there has been very little work on the design and development of computational models that can accurately generate such models.

This paper describes work that has been done to that end, on analyzing such relationships in two separate ant colonies. We discuss methods of observation and describe methodologies and strategies adopted in understanding and simulating the data gathered. We also provide an example of the data accompanied by screenshots of an application that was developed specifically for this purpose. In conclusion, we see how this work is unique in designing systems specifically targeting this particular domain.

## 2. Observing the Swarm

The primary task involved in any such experiment is the actual observation of the ant colony of interest. This task is usually considered to be fairly trivial, but in actuality, it is not.

Our primary assumption in this research was that visualization of colonies of interest in a traditional manner or mode would largely suffice to provide us with the information that is necessary in generating models of interest. While this is still true, we realized over the course of our research that unlike a lot of other insect colony research, predator-prey and friend-foe research involves clusters of insects huddling together.

In the course of analysis, this becomes a problem owing to the difficulty involved in separating and following a particular insect of interest. To this end, we came up with a particular solution that would be useful for such analysis in the future.

### 2.1. Multiple Cameras and Extrusion

While it might seem like an obvious stratagem, it is not usual to find multiple cameras tracking insect colonies. Usually, there is just a single camera positioned over the insect colony of interest, capturing information. However, we found out that two strategically placed cameras helped provide two different perspectives, and points of conflict could be resolved a lot more easily using an overlay model to isolate the areas of interest.

However, we also discovered another methodology of suitably isolating objects of interest, through a process called *extrusion*. In this method, we overlay the regions of interest with a grid matrix with three dimensional density mappings, to find extruded regions of any three dimensional object.

Essentially, this would mean that the floor of the ant habitat would be seen as a plain grid, while the ant itself would be shown as a density mound over the otherwise uniform grid, stretching the grid. This is done by converting the two dimensional image into a set of points 3-dimensional and spatial in nature.

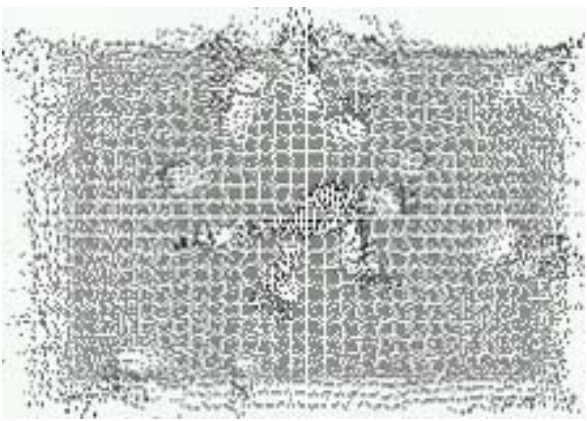

**Fig. 1: Extrusion operation on an ant colony**

Using a time division analysis, this procedure helped eliminate several ambiguous situations where determining the object of interest was non-trivial. It must also be noted that while this is not a solution to having multiple perspective cameras during the course of observation, this methodology did help ease our problems in clustering and other issues that come with observing such entities.

## 3. Studying the Swarm

For our experiment, we used two colonies of carpenter ants, one of which was thriving. We used a standard video capture device to record the data and exported this data into Apple Quick Time format, since our existing systems were capable of processing and analyzing this format for patterns of interest.

### 3.1 Architecture and Environment

For studying the colonies of interest, we used an existing setup at the BORG Laboratory at Georgia Tech for observation and analysis. Other than the manual observation of the insect movement within the colony, we also used automated tracking and analysis system of the insect colonies for data gathering purposes.

The accompanying diagram shows the architecture of the environment that was used to this end:

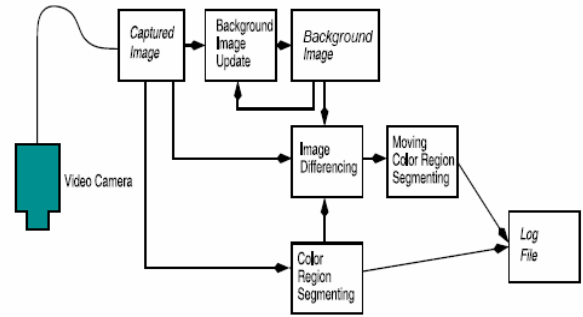

**Fig. 2: Automated Observation Setup [1]**

### 3.1 Experimental Scenarios

Once the system had been setup, the experiment itself consisted of the following case scenarios:

1. Observation of a foreign entity – This was be a non-larval entity that was entirely new to the colony, observed under various conditions (such as forage), interaction (for example, in the case of ants, with minor workers, media workers and major workers) and other relevant aspects, including friend/foe identification.

2. Observation of a larval foreign entity – This was similar to the previous case, except that the entity under observation was a larval foreign entity, primarily to observe whether the larvae was accepted or rejected by the colony and its reactions to various tasks and interactions.

3. Observation of a larval local entity – This was the case where the object of interest was a local, transplanted entity which was observed in its interactions. We also took this opportunity to see how the learning of the larvae actually happens, coupled with other attributes.

4. Observation of a mature foreign entity – This was the case where the object of interest was a foreign, transplanted entity which was observed in its interactions. In our case, we used ants from across the colonies as well as army ants, since our colonies of interest were carpenter ants.

5. Interactions between combinations of the above – This consisted of observing all of the above under various conditions to see how predator-prey and friend-foe identification takes place, as well as to observe any interactions and learning related actions that may take place.

## 4. Data Collection and Analysis

Once the preliminary setup is complete, the next step is to appropriately model and analyze the data for patterns and compare these with existing and known mechanisms, within social insect colonies as well as without. The data would then need to be analyzed for patterns in interactions, identification methods, learning and any other attributes that maybe observed/needed. These patterns would then need to be generalized to accurately predict the behavior of elements within the colony.

In our experiment, we realized that a standard 3 minute video comprising of about 5,400 frames usually tended to provide too many data points that would prove hard to process in real-time. Therefore, we adopted a strategy of skipping every one in three frames to minimize our dataset, to perform real-time analysis.

It must be noted that if real-time analysis is not the goal, then there is no particular need for skipping the frames. It has been our experience, however, that this has no effect on the actual results obtained, in our experiments.

## 5. Data Processing: An Overview

Once the videos have been captured with the appropriate resolution and frame rate, it would need to be analyzed for particulars of interest. The following series of steps describe in details the methodologies that we had adopted to this end.

### 5.1. Preliminary Analysis

The first step in analyzing the data is to standardize the frame rate, and isolate specific sequences of interest. Once this is done, we would have a series of frames, each depicting the object(s) and scene(s) of interest. In our case, these were done at a lower frame rate to help speed up the processing time; however it must be noted that a higher accuracy will be achieved with a higher frame rate. But significantly higher frame rates will deem it impossible to perform the data analysis at a higher resolution, or will entail significant processing power.

Once extracted, these frames are then reconstructed separately to form movies with lower frame rates, to help us in our analysis. When that has been completed, we proceed to perform other operations upon the frames that were used to reconstruct the newer video with a lower frame rate.

The objective of the first set of these operations is two fold: locate the entities of interest and mark trails.

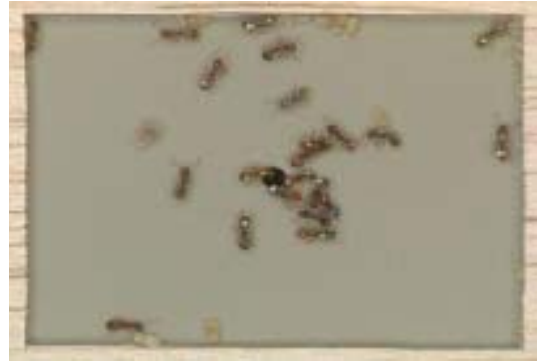

**Fig. 3: Sample Data Frame**

### 5.2. Extrusion Operation and Analysis

When the frames have been extracted and the movie reconstructed at the desired frame rate, our next step is to perform what we call an *extrusion operation* on a frame-by-frame basis. The idea behind the extrusion operation is to primarily convert the flat image of a frame into spatial data points. This is accomplished by overlaying a grid over the flat image, and stretching the grid across regions of varying spatial densities.

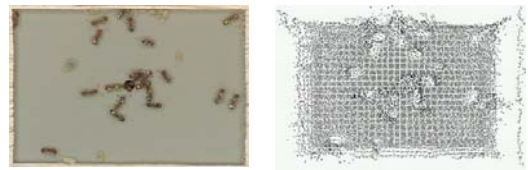

**Fig. 4: Normal Frame versus Extruded Frame**

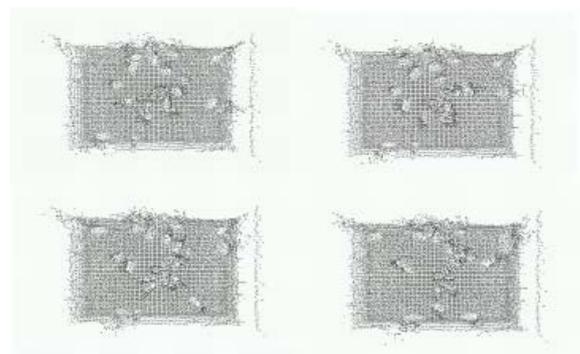

**Fig. 5: Extruded sequence of frames**

Once the extrusion operation has been completed on all the frames, a new movie is reconstructed consisting solely of extruded data. These are then analyzed for regions where there is movement and regions where there are not. These helped us look for clusters and other interesting data points during the course of our analysis.

### 5.3. Locating Entities of Interest

With extrusion complete, we then proceeded to perform a base setup of locating objects of interest, on both the actual frames and the extruded frames. These were then compared with one another to establish definite zones of interest.

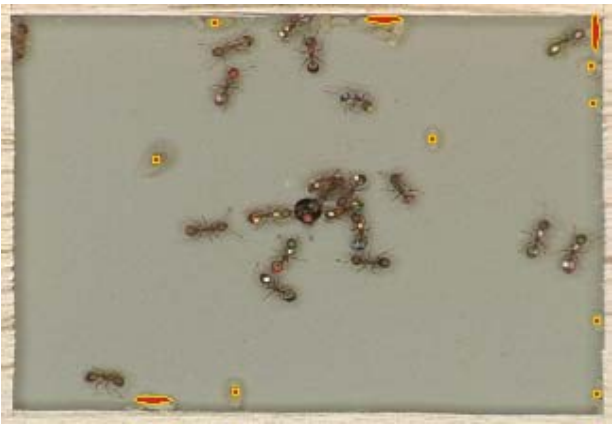

**Fig. 6: Zones of interest in regular frames**

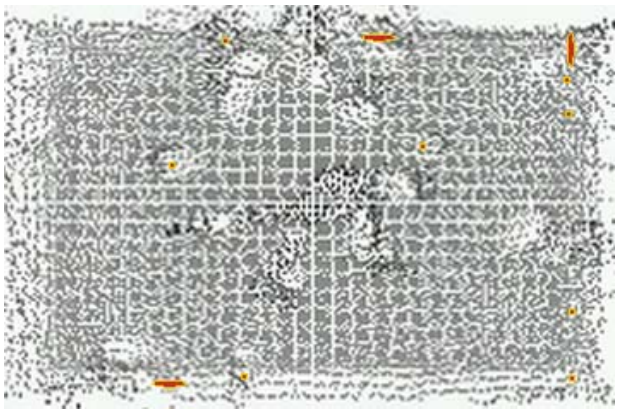

**Fig. 7: Zones of interest in extruded frames**

In this particular case, we observe friend interaction among larvae belonging to the same colony. Zones of interest were defined as regions where there were larvae, determined by regions that had a higher density than the surface (indicating the presence of an object) but having absolutely no motion. It is interesting to note that there were instances when stationary ants were also marked as being entities, owing to their absolute lack of motion.

### 5.4. Generating Trails

Once these regions of interest were defined, our next step was to generate trails of the objects of interest. For this, we used an existing tool developed at the BORG Lab, called Team View [2], to track two ants of interest.

It should be noted that higher the number of tracks recorded, better the accuracy and precision. For a smaller colony, however, two tracks usually suffice.

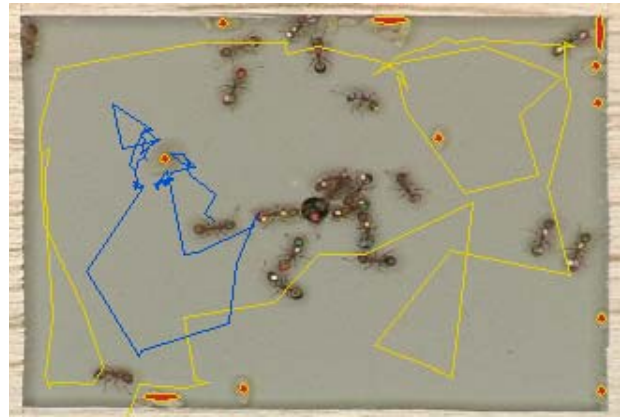

**Fig. 8: Track of ants of interest on a regular frame**

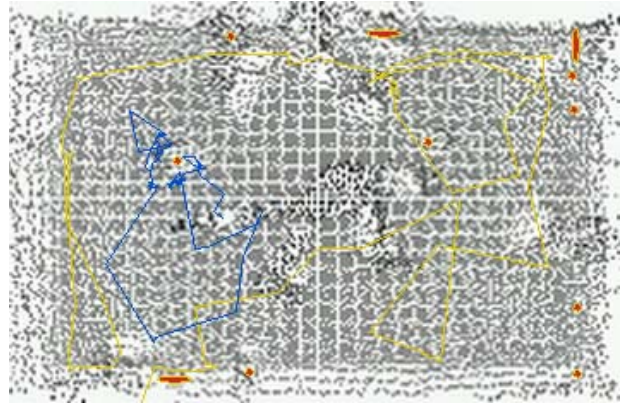

**Fig. 9: Track of ants of interest on an extruded frame**

In this particular we recorded one track of an ant that had interaction with the larvae and another ant that did not have any such interaction. Once again, these steps were repeated on the frames that were extruded, as well, to provide for reference frames.

### 5.5. K-Nearest Neighbor

Once the tracks were established, we also noted stationary regions where the ant that had interactions with the larvae had stopped, and performed correlated these with existing data on larvae. Thus, we had entity zones defined on the basis of similarities in regions (i.e. interest zones) and similarity in movements. A K-Nearest-Neighbor analysis was done on all such data points, to indicate either larvae or a potential track and position for larvae.

These were performed on the normal frame first, and compared with the extruded frame. The data was correlated to be summative (i.e. add areas that did not match) and not differential (i.e. remove areas that did not match). This was so that we did not miss out on any potential data points.

This was the final step in our pre-processing of the data. It must be noted that for better accuracy, a KNN operation could also be performed upon ants of interest, which would yield more information.

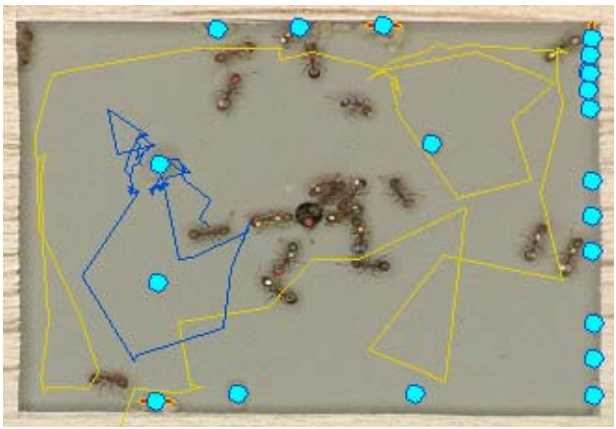

**Fig. 10: KNN of larvae on a regular frame**

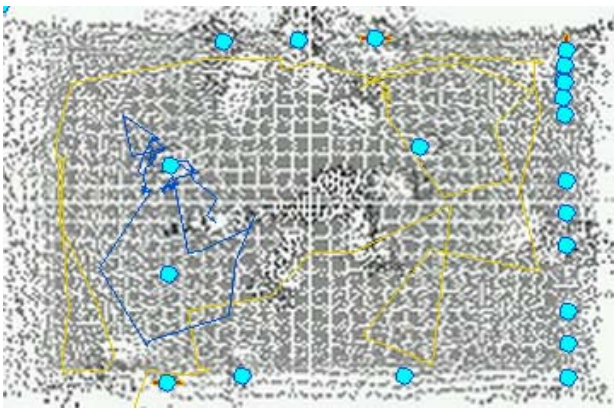

**Fig. 11: KNN of larvae on an extruded frame**

### 5.6. Movement Probability Distributions

When the preprocessing of the data was completed, our next step was to analyze the movement of the ants and determine a probability distribution with respect to each other, location of the objects of interest and any other parameter. The last is for any interaction that we are unable to determine/process, so that we may account for its presence in our prediction and simulation.

Each of these were denoted by a color code, where red indicated an ant interaction, green indicated larvae interaction and blue indicated any other interaction that we had not accounted for previously.

It must be noted that these are probability distributions, and we assumed a uniform zone of influence for each of these data points. However, these may vary based on the colony in question, and perhaps even from one ant to another. Increased precision may require further tuning of the system to take these into account.

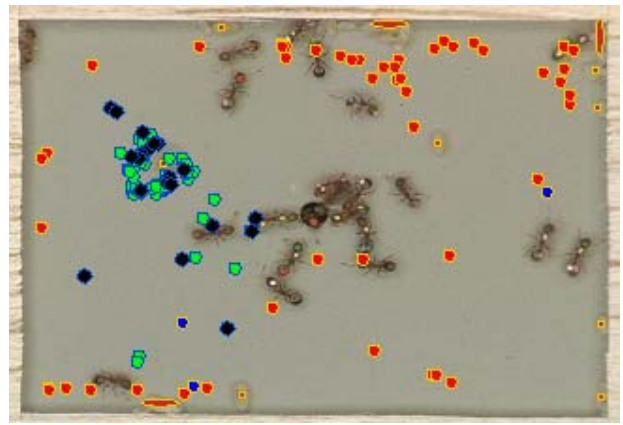

**Fig. 12: Movement probability distribution zones**

*Legend*

- Red: Ant-Ant Interaction
- Green: Ant-Larvae Interaction
- Blue: Other Interaction

### 5.7. Markov-Chain Monte-Carlo Analysis of P

Once these tracks were established, our next task was to reassign the probability values for each of the three conditions.

What this means is that, at any given point in our colony, we would need to determine the probability of an ant being under the influence of another ant, influence of a larvae or the influence of something else.

While these were the only three conditions that we considered, it is possible to perform a more detailed analysis depending on the extent to which one may wish to focus.

This is a challenging task, because there is no good way to determine this a priori. However, what we could do was to assume random points on the frame of interest assume ants in those positions and calculate the probability values. However, this would entail us to have an initialization sequence, where the whole frame would be analyzed from a particular frame.

Therefore, we assumed that we did not care *how* the ant arrived at a particular frame, only that it was already there and that we would analyze its future positions. So, we considered each movement in a frame to be discrete (i.e. a state). Thus, movement from one state to another is not random; however present state could be randomly initialized.

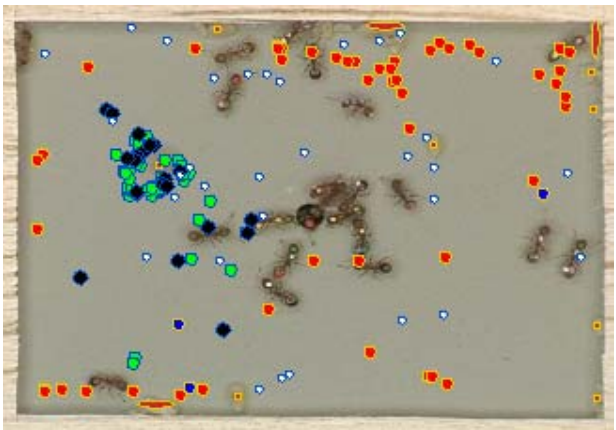

**Fig. 13: Random ant positions, indicated by the white dots**

For each of these random states that are assigned, we compare them with identical or similar ant states and look for patterns. Based on these, the probabilities of that state and possible transition states are established, as a function of three probabilities:

- P (ant)
- P (entity)
- P (other)

Thus, we've the following series of steps that are performed:

- Random state assignment
- Comparison with similar ant states
- Fine Tuning of P (ant, entity, other)

### 5.8. Back Propagation

Once the MCMC analysis of P is complete, the the probabilities have been established (independently) for each of the random states. However, the next challenge is to classify these independent probabilities for each of the random states.

This depicts a very classic case of perceptron classification problem, which can be quite easily solved by using a weighted back propagation network.

| State | K (State) | W0 | W1 | W2 |
|-------|-----------|------|------|------|
| *1* | *k1* | *w0 (k1)* | *w1 (k1)* | *w2 (k1)* |
| *2* | *k2* | *w0 (k2)* | *w1 (k2)* | *w2 (k2)* |
| . | . | . | . | . |
| . | . | . | . | . |
| . | . | . | . | . |
| *N* | *kn* | *w0 (kn)* | *w1 (kn)* | *w2 (kn)* |

In the above table, the states {1, 2 … N} depict each of the random states that have been assigned, giving us N input cases. The weights W0, W1 and W2 present us with the three classifications of the probabilities for each of the state, given by the function of K, which is defined by the probabilities of each random state.

This operation is performed until the weights reach acceptable levels of probability, or until one or more of the weights approach triviality. Hence, we iterate through until results match within acceptable levels of probability for each of the random states.

### 5.9. Random Walk Simulation

Once we have obtained a suitable set of weights for each probabilistic influence given a random state, we perform a random walk on the states, with respect to the position of the ants.

Once again, random states are chosen and once a random state is assigned, choose all possible paths from that state. This means that given a particular state, we look at all possible states that the ant could go to, based on the previous MCMC model that was obtained (after performing a weighted analysis from the model, using the weights obtained in the previous step).

Therefore, we have the initial random state from which several forks spring out, and each of these forks has several other forks and so on. This is continued till a state is reached where the probability of its existence is below acceptable limits (we compare the forked values based on the original MCMC values that were obtained).

Once such a state has been reached, we backtrack and eliminate all low probability shoots leading to that state.

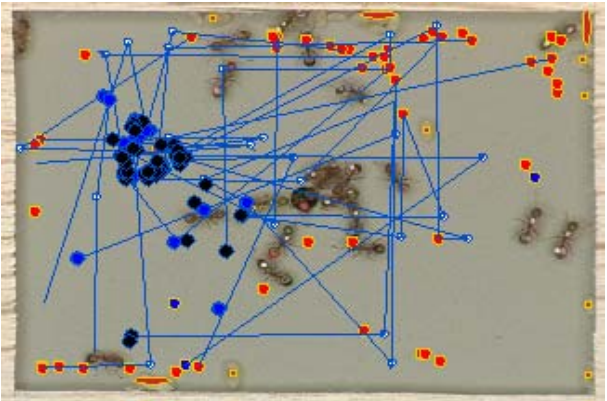

**Fig. 14: State expansion nodes at a given point in time**

We allow two kinds of random walks allowed – one that is System Defined & one that is User Defined. This also entails the user to interfere and introduce any changes or corrections into the learning, as and when it occurs.

Once this is done, we have probable future states for all positions given a random position. We iterate this process on all regions of each frame, in order to provide us with data for any possible position.

The end result is that given a particular position, a probable set of future states can be attained.

### 5.10. Track Projection and Simulation

The above procedure is repeated for each of the frames, until we have the weighted probabilities for each frame, given the state of an ant. Thus, we have a probable set of future states for any ant (new or existing) on any position in any frame.

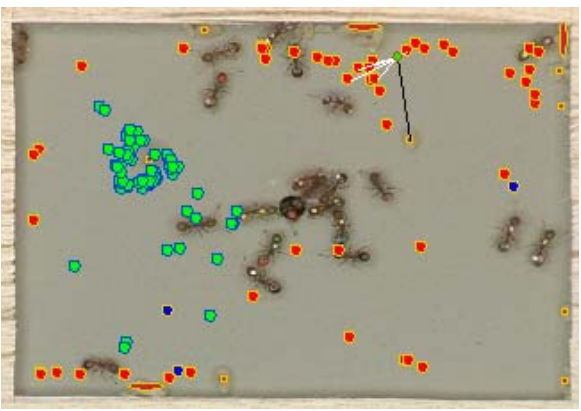

**Fig. 15: Probable states given a position**
*(White: non-entity induced; Black: entity induced)*

The simulation performs this operation on all the frames, and repeats the classification of the probabilities between and across frames (i.e. within each frame and between all the frames). This also helps the system to provide a more refined set of probabilities for a given state in each frame.

This also ensures that the probabilities defined by the interactions not involving ants and entities are minimized. Therefore, it has been observed that states with P (other) usually tend to get eliminated. This is emergent behavior.

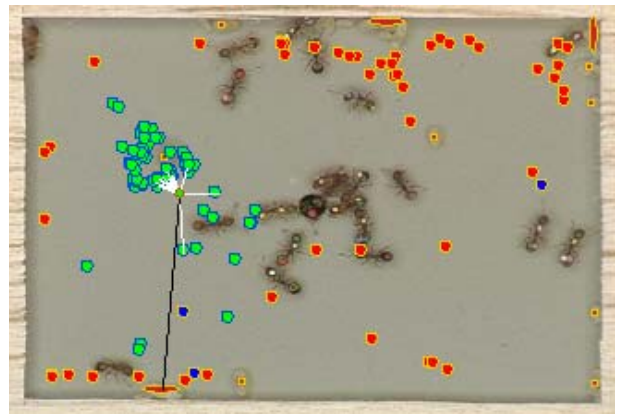

**Fig. 16: Probable state prediction: States and Positions with P (other) depicted by blue dots is minimized to just 3 instances in this particular frame.**

Thus, we can predict a probable set of future positions, given a particular position.

## 6. Live Prediction: Final Simulation

Once we can predict a probable set of states given a state, we attempt simulating the position of an existing ant on the frames, given its position.

This is done without any pre-defined tracks or interaction probabilities, except for zones depicting the positions of entities of interest.

In the following couple of figures, we see our system depicting the possible future states that an ant could take. Both the depictions are of the same frame, but of two different ants.

The first ant is likely to be influenced by the entity (i.e. larvae, indicated by the black line from the ant's position to the larvae that may possibly influence); however, the second ant has no entity influence whatsoever (i.e. only white lines depicting future paths influenced by other ants).

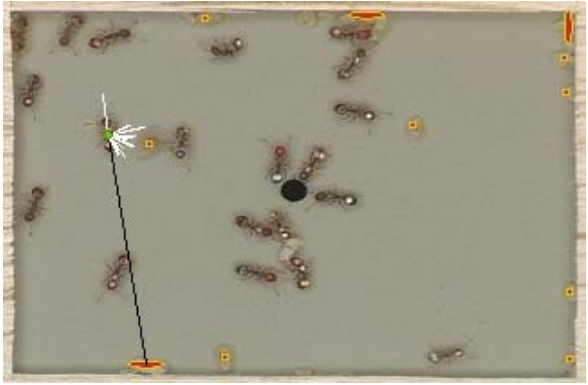

**Fig. 17 (a): Live Prediction of an ant's probable future positions**
*(White: non-entity induced; Black: entity induced)*

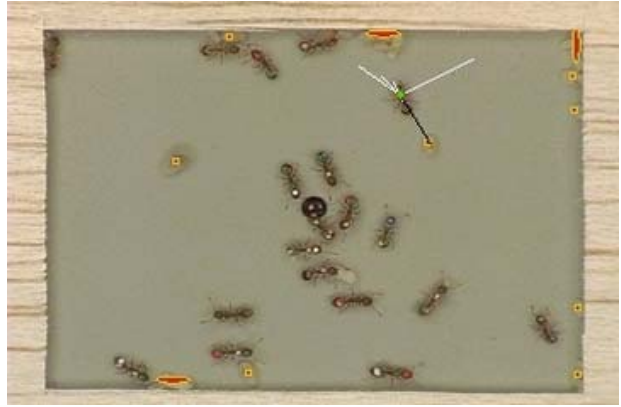

**Fig. 18 (b): Prediction of ant's future states: frame 2**

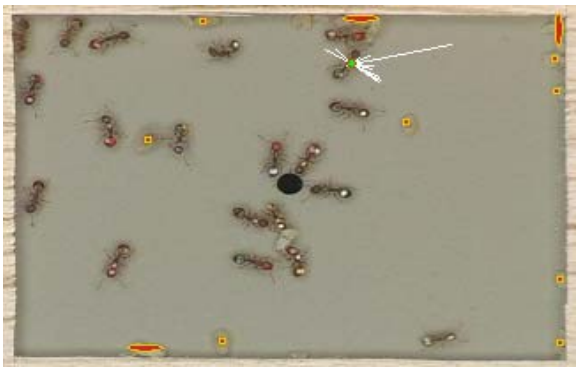

**Fig. 17 (b): Live Prediction of another ant's future positions, on the same frame. If you notice, this ant has no entity induced states in its future path.**
*(White: non-entity induced; Black: entity induced)*

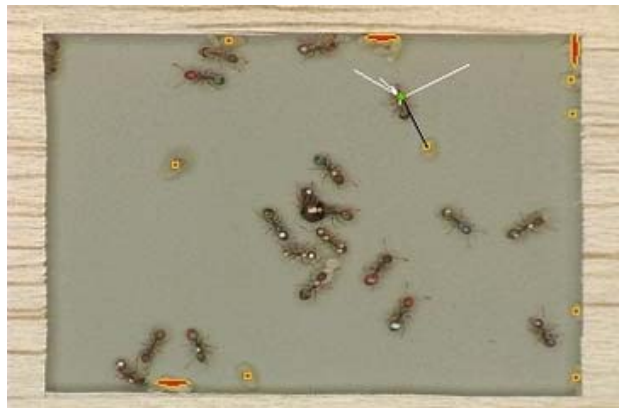

**Fig. 18 (c): Prediction of ant's future states: frame 3**

The following figures demonstrate the sample live prediction of an ant's probably future states, as it moves across over multiple frames.

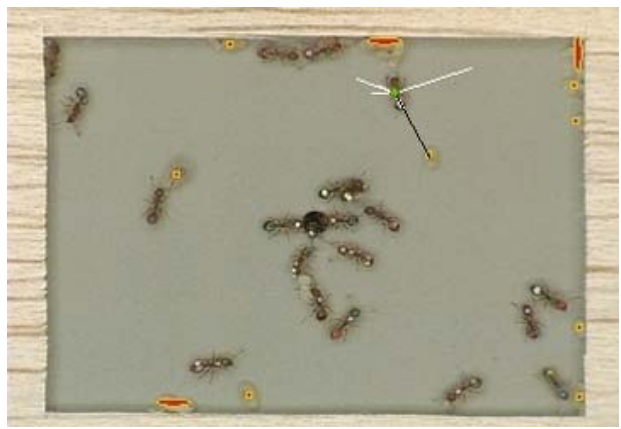

**Fig. 18 (d): Prediction of ant's future states: frame 4**

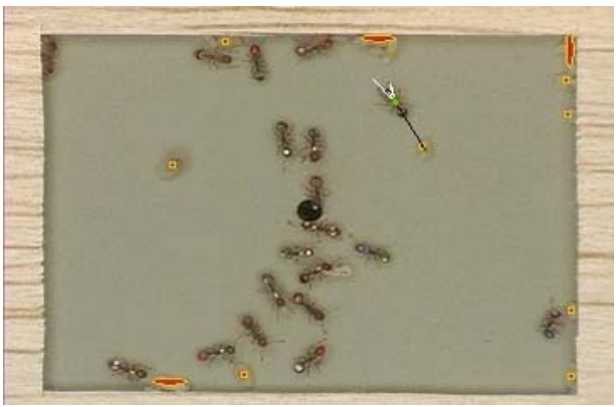

**Fig. 18 (a): Prediction of ant's future states: frame 1**

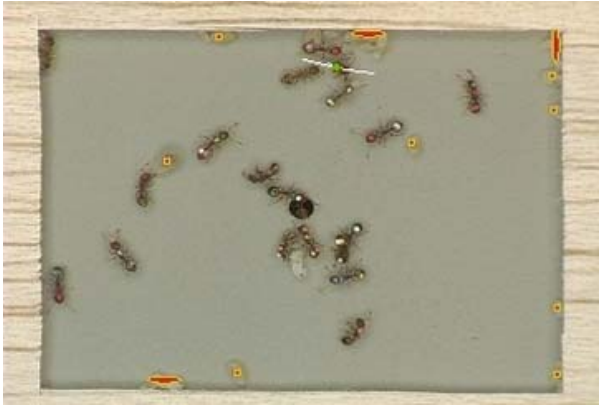

**Fig. 18 (e): Prediction of ant's future states: frame 5**

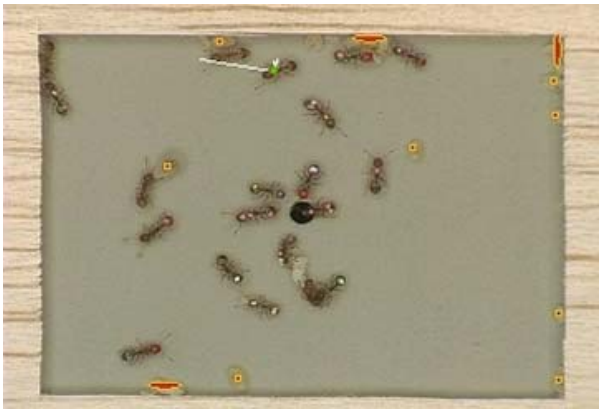

**Fig. 18 (f): Prediction of ant's future states: frame 6**

As you can see, the above sequence shows an ant being tracked and its future states and positions (accurately) predicted across multiple frames.

## 7. Next Steps

This work has only been done on ant colonies under controlled situations. It would be a challenging task to generalize and extend this work to larger colonies, and ultimately, to swarms. We would like to eventually be able to run swarm simulations to accommodate group theories of collectives, to help predict similar behaviors in swarms.

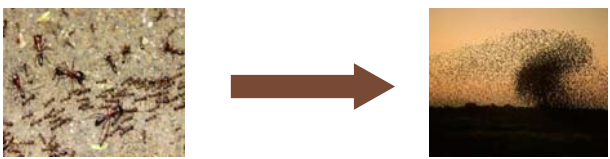

## 8. BORG Lab Applications

The successful simulation and prediction of ant states can be extended further to live robotic simulation, using GNATs to simulate friend-foe and predator-prey behavior. Ants often use these techniques to solve problems under various situations. Similarly, these simulations may help design GNATs that can solve problems better.

These behavior simulations can also be used to model friend-foe and predator-prey like algorithms in robot soccer, and help train them better.

## 9. Applications

There are several applications to the simulations that we have developed. Primarily, these provide biologists with information and models to accurately gauge behavioral controllers.

These can also be used in algorithmic domains for analogous situations, adapting the techniques and methods learnt to situations at hand.

Such adaptations can also be extended to strategies that are learnt in real-life friend-foe and predator-prey scenarios, such as wars and military planning (e.g. UAV reconnaissance).

## 10. Conclusion & Future Work

We were able to develop a system that was successfully able to predict ants under certain conditions. We also designed a framework for building future systems that can help us in similar predictions.

For future work, this framework can be extended and improved upon in several ways, in terms of optimizing capture and customizing the system for a specific task at hand. We could also extend this system to other insect colonies and swarms, such as bees, locusts etc. as well as other species that demonstrate a collective behavior (e.g. fish).